\documentclass[runningheads]{llncs}
\usepackage{graphicx}
\usepackage{tikz} 
\usepackage[normalem]{ulem}
\usepackage{amsmath}
\usepackage{subcaption}

\usepackage{amsfonts}
\usepackage{eucal}

\begin{document}
\title{DOT-VAE: Disentangling One Factor at a Time}
%
%
\author{Vaishnavi Patil \and
Matthew Evanusa \and
Joseph JaJa}
\authorrunning{Patil et al.}
%
\institute{University of Maryland, College Park, MD 20740, USA \\
\email{\{vspatil, mevanusa, josephj\}@umd.edu}\\
}
\maketitle              
\begin{abstract}
        As we enter the era of machine learning characterized by an overabundance of data, discovery, organization, and interpretation of the data in an \textit{unsupervised} manner becomes a critical need. One promising approach to this endeavour is the problem of \textit{Disentanglement}, which aims at learning the underlying generative latent factors, called the factors of variation, of the data and encoding them in disjoint latent representations. Recent advances have made efforts to solve this problem for synthetic datasets generated by a fixed set of independent factors of variation. Here, we propose to extend this to real-world datasets with a countable number of factors of variations. We propose a novel framework which augments the latent space of a Variational Autoencoders with a disentangled space and is trained using a Wake-Sleep-inspired two-step algorithm for unsupervised disentanglement. Our network learns to disentangle interpretable, independent factors from the data ``one at a time", and encode it in different dimensions of the disentangled latent space, while making no prior assumptions about the number of factors or their joint distribution. We demonstrate its quantitative and qualitative effectiveness by evaluating the latent representations learned on two synthetic benchmark datasets; DSprites and 3DShapes and on a real datasets CelebA.

\end{abstract}
\section{Introduction}
Deep learning models, which are now widely adopted across multiple Artificial Intelligence tasks ranging from vision to music generation to game playing \cite{kri,oord,minh}, owe their success to their ability to learn representations from the data rather than requiring hand-crafted features.  However, this self-learning of abstract representations comes at the known cost of the resulting representations being cryptic and inscrutable to human observers \cite{shortcut}. A more comprehensive representation of the data where the essential indivisible, semantic concepts are encoded in structurally disentangled parts could lead to successful domain adaptation and transfer learning \cite{bengio} and facilitate robust downstream learning more effectively \cite{suter}. Learning these latent representations from the data alone without the need of laborious labeling by human observers constitutes the problem of \textit{Unsupervised Disentanglement}.  In this work, we attempt to address the problem of unsupervised disentanglement via a novel Variational Autoencoder based framework and training algorithm.




Though there is no commonly accepted formalized notion of disentanglement or validation metrics \cite{def}, recent works have characterized disentangled representations, based on natural intuition. This intuition by \cite{bengio} states that a disentangled representation is a representation of the data which encodes each \textit{factor of variation} in disjoint sets of the latent representation. \cite{loca} state further that a change in a single factor of variation produces a change in only a subset of the learned latent representation which corresponds to that factor. Here, a factor of variation is an abstract human defined concept that assumes different values for different examples in the dataset. This intuition is closely related to the independent mechanisms assumption \cite{causal} which renders the informative factors as components of a causal mechanism. This assumption allows interventions on one factor without affecting the other factors or the representations corresponding to the other factors and thus can be independently controlled. In our work we use independent interventions on the learned disentangled representations, which encode the different factors, to generate samples and restrict the differences in corresponding representations, of the data and the sample, pertaining to that factor. This process of using interventions and generating new samples resembles the sleep phase of the wake-sleep algorithm. 

Most current Variational Autoencoder (VAE) based state-of-the-art (SOTA) make the implicit assumption that there are a fixed number of independent factors, common for all the data points in the dataset. However in real datasets, in addition to the independent factors common to all points in the dataset, there might also be some correlated, noisy factors pertinent to only certain data points. While the approaches based on Generative Adversarial Networks (GAN) do not make an assumption, they learn only a subset of the disentangled factors, whose number is heuristically chosen. We believe however that one of the main goals of disentanglement is to glean insights to the data, and the number of factors of variation is generally one that we do not have access to.  To this end, our method augments the entangled latent space of a VAE with a disentangled latent code, and iteratively encodes each factor, common to all the data points, in a single disentangled code using interventions. This process allows our model to learn any number of factors, without prior knowledge or "hardcoding", thus making it better suited for real datasets. 
\subsection{Main Contributions}
Our contributions in the proposed work are:
\vspace{-2mm}
\begin{itemize}
\item We introduce a novel, completely unsupervised method for solving disentanglement, which offers the mode-covering properties of a VAE along with the interpretability of the factors afforded by the GANs, to better encode the factors of variations in the disentangled code, while encoding the other informative factors in the entangled representations. 
\item Our proposed model is the first unsupervised method that is capable of learning an arbitrary number of latent factors via iterative unsupervised interventions in the latent space.
\item We test and evaluate our algorithm on two standard datasets and across multiple quantitative SOTA metrics and qualitatively on one dataset. Our qualitative empirical results on synthetic datasets show that our model successfully disentangles independent factors. Across all quantitative metrics, our model generally outperforms existing methods for unsupervised disentanglement.
\end{itemize}
\footnotetext{Code available at https://github.com/DOTFactor/DOTFactor} 
\section{Disentangling One Factor at a Time using Interventions}
We base our framework on the VAE which assumes that data $x$ is generated from a set of latent features $z \in \mathbb{R}^d$ with a prior $p(z)$. A generator $p_\theta(x|z)$ maps the latent features to the high-dimensional data $x$. The generator, modeled as a neural network, is trained to maximize the total log-likelihood of the data, $\log p_\theta(X)$. However, due to the intractability of calculating the exact log-likelihood, the VAE instead maximizes an evidence lower-bound (ELBO) using an approximate posterior distribution $q_\phi (z|x)$ modeled by an encoder neural network. For given data, this encoder projects the data to a lower-dimensional representation $z$, such that the data can be reconstructed by the generator given only the representation. Without any structural constraints, the dimensions of the inferred latent representation $z$ are informative but entangled.  
\textbf{Disjoint Latent Sets:}
To encode the factors of variation in an interpretable way, following \cite{hu} we augment the unstructured variables $z$ with a set of structured variables $c = \{c_1, c_2, \cdots, c_K\}$ each of which is tasked with disentangling an independent semantic attribute of the data. We train our network to systematically discern the meaningful latent factors shared across the dataset into $c$, from the entangled representation $z$, both of which are important to maximize the log-likelihood of the data. However, it is only the most informative, common factors of variation encoded in $c$ that we are interested in, as the remaining factors in $z$ are confounded and contain the 'noise' in the dataset, i.e., features that only a few data samples contain.

The generator is now conditioned both on the disentangled latent codes and the entangled latent space $(c,z)$ and describes a causal mechanism \cite{suter}, where each causal component $c_k$ is independently controllable and a change in a particular index $k$ has no effect on any other index $c_{j} (j\neq k)$. Manipulating each $c_k$ should correspond to a distinct, semantic change, corresponding to the factor encoded, in the generated sample, without entangling with changes effected by the other factors or with the entangled code $z$. To this end, we perform \textit{interventions} as proposed in \cite{suter} (described in detail below) that go into the latent code, change a single dimension $c_k$ of the disentangled code $c$ while keeping the other dimensions the same. We generate data from this intervened latent and then constrain the model to reconstruct the exact change.

To encourage the generator to make distinct changes for each $c_k$ that is manipulated during interventions, we re-encode the data to recover the manipulated latent representation that was used to generate the data. Thus if the code $c_i$ was manipulated, while keeping the rest $c_{j} (j\neq i)$ and $z$ unchanged, the encoding of the corresponding generated data must reflect a change only in the manipulated code $c_i$.

\begin{figure}[t]
\begin{center}
\includegraphics[width=0.7\textwidth]{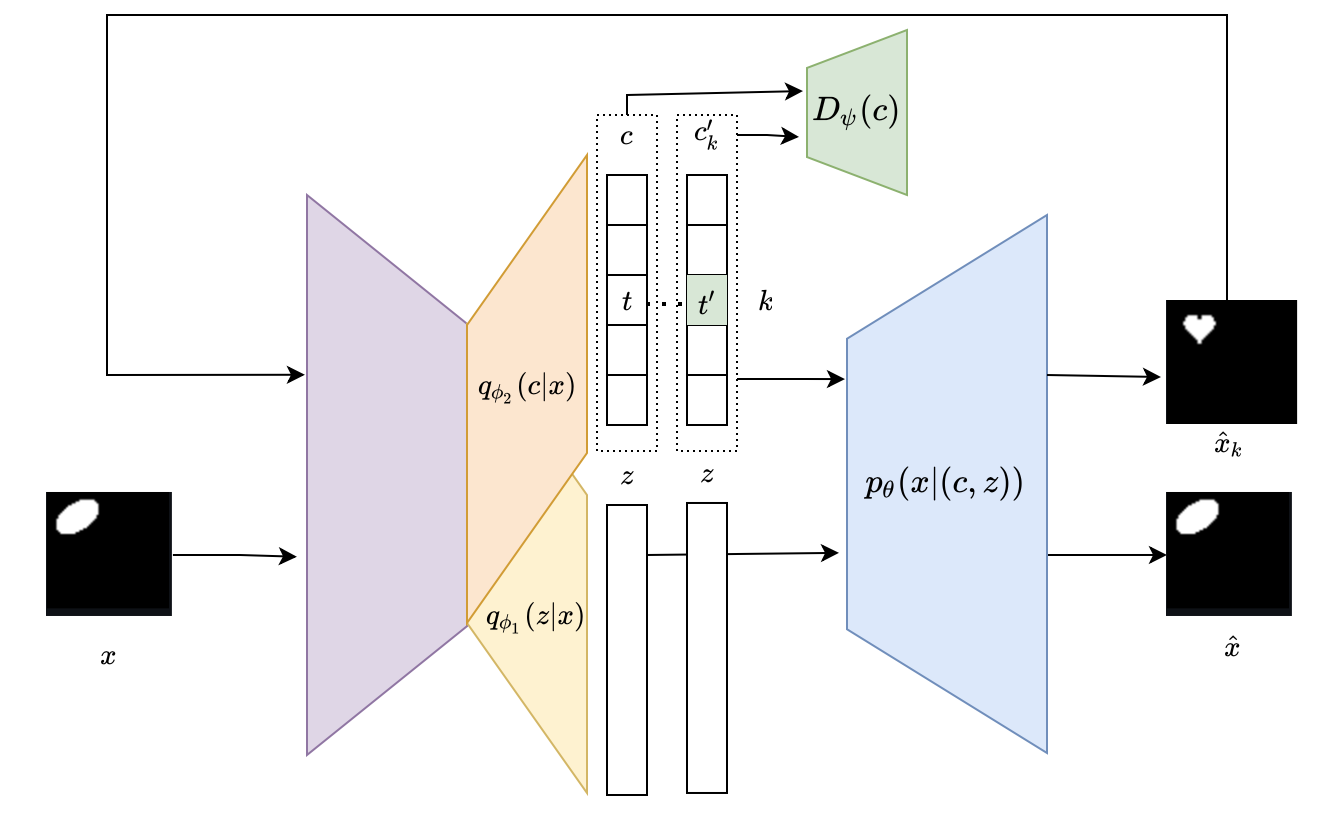}
\caption{The full architecture of Dot-VAE. Input data is fed into an encoder, which projects the data into two disjoint latent spaces, a disentangled $c$ and entangled $z$. \textit{Interventions} change the value $t$ of a single disentangled latent at index $k$ to a new value $t^{\prime}$ while keeping the other latents unchanged to get the intervened representation $(c^{\prime}_k, z)$. The adversarial network ensures that the distribution of the real representations $c$ and the intervened representations $c^{\prime}_k$ is close to each other. The decoder maps the intervened latents to generated data $\hat{x}_k$. This new generated data is passed back through the encoder to train the generator to make distinct and noticeable changes.} \label{fig1}
\end{center}
\end{figure}

\noindent \textbf{Adversarial Latent Network:} Moreover, in order to ensure that interpretable factors are encoded in the disentangled code $c$, the changes effected by each $c_k$ must correspond to a semantic change in the generated data corresponding to a change which could be effected by a factor of variation. For this, the data generated from the disentangled code manipulation during interventions must lie in the true data distribution. To ensure this we train a discriminator, like in \cite{factor}, in the latent space such that the latent representations after manipulations continue to lie in the distribution of the encodings of the true data. The encoder is trained to reduce the distance between the distribution of the representations, used by the generator to reconstruct the data, and the distribution of the representations after they have been intervened on. This in effect helps us train a generative model which can be conditioned to generate realistic, new samples with any desired values for the different latent factor. Figure \ref{fig1} shows the overall framework which is trained with a two step algorithm inspired by the wake-sleep algorithm.


\textbf{Model Structure and Learning: }
In this section we will detail the training method, and how to combine the architectural components described earlier. We augment the latent space $z$ to include a disentangled set of latent codes $c$ with prior $p(c) = \prod_{i=1}^K p(c_i)$, where $p(c_i) \sim \mathcal{N}(0,1)$. The encoder $q_\phi (c,z|x)$ is split into two to model the distributions of the two latent variables as $q_{\phi_1}(z|x)$ and $q_{\phi_2}(c|x)$ respectively. Under this augmented space the evidence lower-bound of the data log-likelihood is as follows:

\begin{equation} \label{eq:1}
    \begin{split}
        \mathcal{L}_{\theta, \phi} = -\mathbb{E}_{q(x)}[\mathbb{E}_{q_{\phi} (c,z|x)} [\log p_\theta (x|c,z)] &+ KL [q_{\phi_1}(z|x) || p(z)] \\ 
    &+ KL [q_{\phi_2}(c|x) || p(c)]]
    \end{split}
\end{equation}
where $q(x)$ is the empirical training data distribution.
The first term in the above objective function minimizes the error of reconstructing the higher-dimensional data from it's lower-dimensional representations. The KL divergence in the second and the third terms ensure that posterior distribution learned by the model is close to the uninformative prior distribution, and can be sampled from for generating new samples. These three terms together create an information bottleneck in the latent space where only the information relevant to recover the data is encoded in the representations. This ensures that all the informative factors of variations are encoded in the representations either in $z$ or in $c$.
\begin{figure}[t]
\begin{subfigure}{0.5\textwidth}
  \centering
  \includegraphics[width=\linewidth]{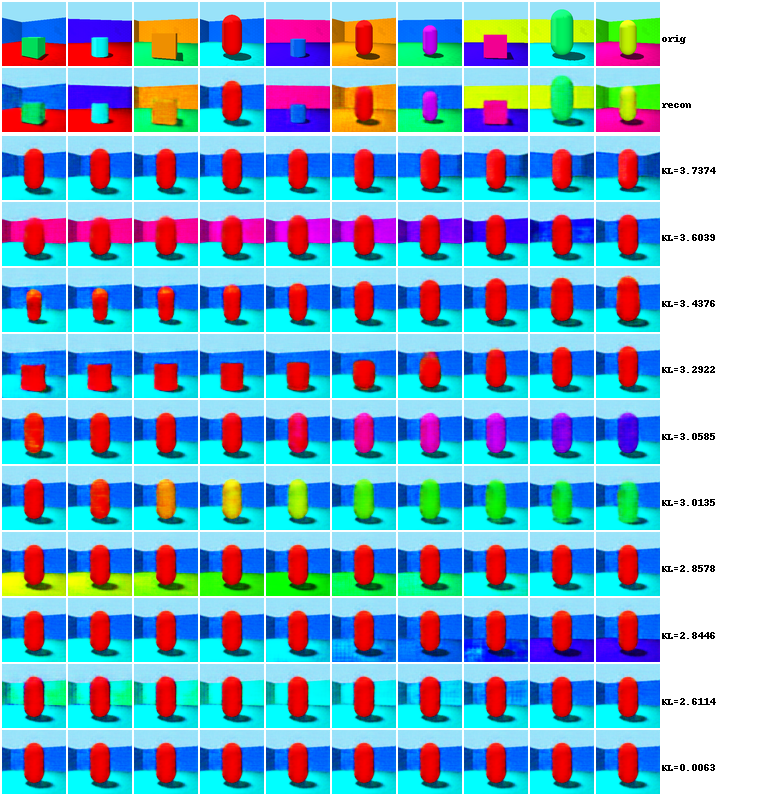}
  \label{fig:sub1}
\end{subfigure}%
\begin{subfigure}{0.5\textwidth}
  \centering
  \includegraphics[width=\linewidth]{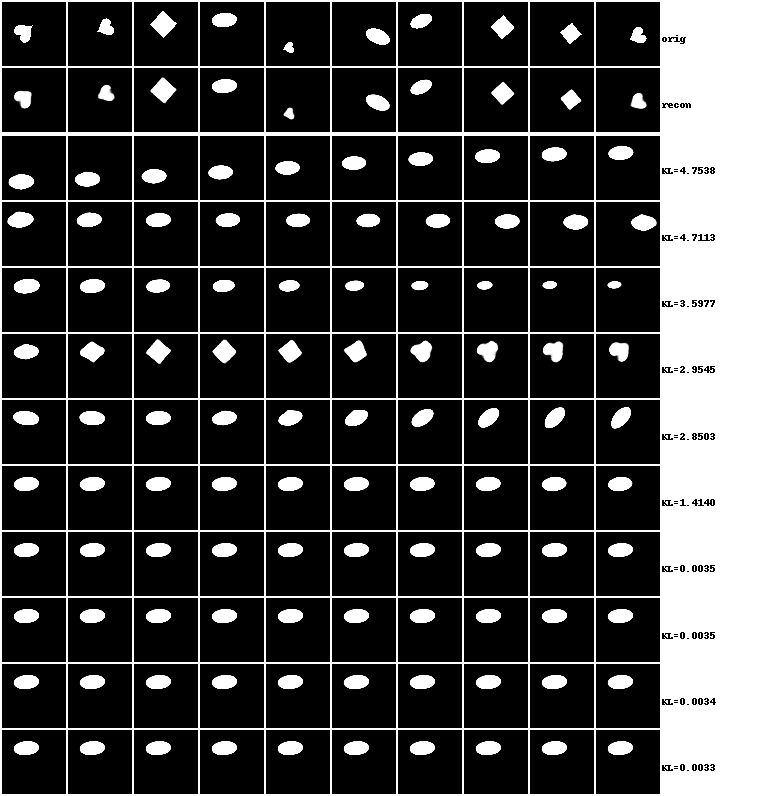}
  \label{fig:sub2}
\end{subfigure}
\caption{Latent traversals for the disentangled code $c$ for 3Dshapes (Left) and Dsprites dataset (Right). Our method disentangles the informative factors and encodes them in the different dimensions of $c$. The top two rows are the original and reconstructed images, respectively. Each row below the second corresponds to a traversal. \textit{3DShapes (Left): } Each row corresponds to a distinct independent factor.  Rows three through ten correspond to orientation, wall hue, size, scale, object hue (two rows), floor color (two rows) respectively.  This covers all of the exact factors of variation.  In the last row, we can see the model discerned it had discovered all the factors of variation, and need not encode anything. \textit{dSprites: (Right) }  Similarly, in dSprites we can also see that the model discerns the factors in rows 3 through 7: x, y coordinate, size, shape, and orientation.}
\label{fig:latent_shapes}
\end{figure}

\subsubsection{Learning One Factor at a Time}
Since we want each disentangled latent code $c_k \in c$ to encode a factor of variation, changes in the value of a latent code $c_k$, while keeping the others the same, should bring about semantic changes in the generated samples corresponding \textit{only} to the encoded factor. The changes should be such that when the original data and the generated data (after interventions) are compared, the encoder must be able to discern (1) the index of the changed latent variable, and (2) by how much it was changed. We enforce these conditions by performing interventions on one code at a time and reconstructing the latents of the interventions, sequentially. At the beginning of training, we start with interventions on the first element of the disentangled space $c_1$. Once the objective in Equation \ref{eq:1} saturates for interventions with $c_1$, we shift the interventions to the next element in the set, $c_2$ along with $c_1$. As the training proceeds, more dimensions of $c$ are intervened upon together and the network learns to encode the confounding information into $z$, while keeping only the disentangled information in $c$. For synthetic datasets with a fixed number of factors of variations, we continue iterating through the elements of $c$ until the entangled part of the representation $z$ encodes no information i.e. $KL(q_{\phi_1} (z|x) || p(z)) = 0$. For real-world datasets we can continue the iteration until we find a desired number of factors. 

\begin{figure}[t]
  \centering
  \begin{subfigure}{0.5\textwidth}

   \includegraphics[width=\textwidth]{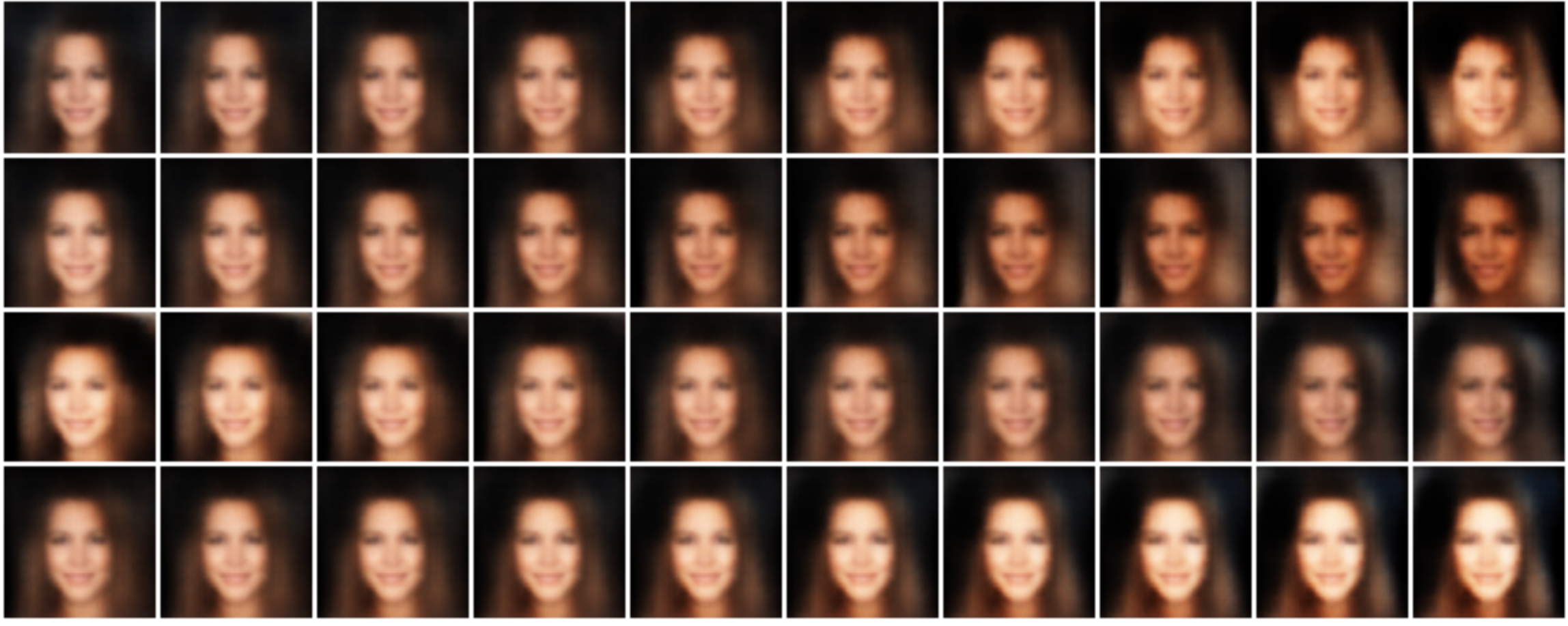}
    \end{subfigure}%
    \begin{subfigure}{0.5\textwidth}
    \centering
    \includegraphics[width=\textwidth]{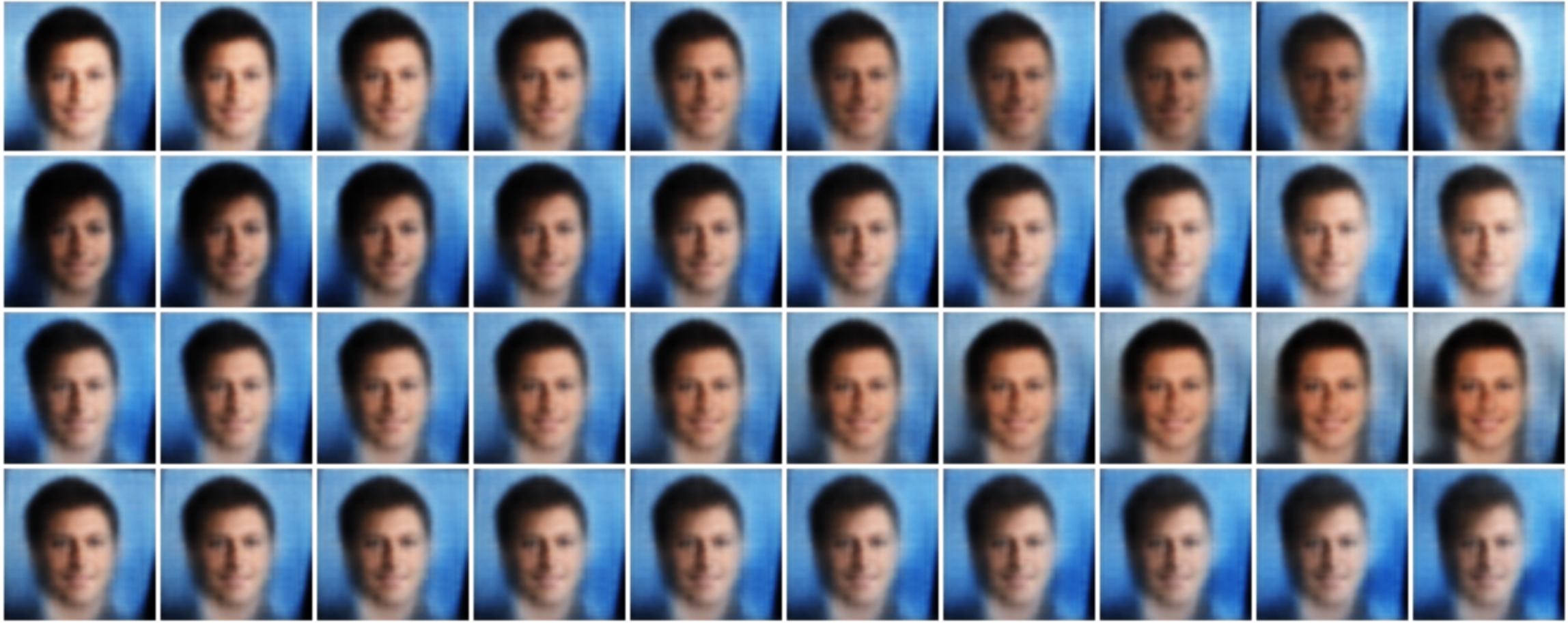}
    \end{subfigure}
\vspace{-2mm}
\caption{Traversals over latent space $c$ for the \textit{CelebA} real-world celebrity dataset. The dataset consists of cropped headshots of various celebrities. Given a seed image, we perform a traversal across various latents.  Our model successfully produces reconstructions that correspond to identifiable factors. Like for synthetic data, some of the latents remain unused; shown are exemplar latents that encoded factors of variation. The model learns many more factors as compared to previous methods such as hair color, skin tone, lighting, etc.} 
\label{fig::celebA}
\end{figure}


Formally, given a batch $B$ of examples from the data distribution $\{x^i\}_{i=1}^B$, which we write as $\{x^i\}^B$ for brevity, we first encode the batch to find their latent representation $\{(c^i, z^i) \sim q(c,z|x^i)\}^B$. We then select an index $k \in [K]$ and proceed to intervene/change the value of the element $c_k$, which we refer to as $\{t^i\}^B$, of the disentangled code, to $\{{t^{\prime}}^i\}^B$ (i.e. to the value of another example $t^l (l \neq i)$ at the same dimension $c_k$), for all the representations in the batch, while keeping the other disentangled codes $c_j (j \ne k)$ and the entangled representations $z$ the same. We change the value only for that code to obtain our intervened latent representation $\{{c^{\prime}_k}^i = ({t^{\prime}}^i, c_j^i (j \neq k))\}^B$. We then use these intervened representations to generate a batch of samples $\{\hat{x}_k^i \sim p_\theta (x|{c^{\prime}_k}^i, z^i)\}^B$. 

The generated samples $\{\hat{x}_k^i\}^B$ must differ from the data examples $\{x^i\}^B$ only in the factor corresponding to one encoded in $c_k$. To ensure this, we re-encode the generated samples to reconstruct the intervened latent representation $\{{\hat{c}_k}^i \sim q_\phi (c,z|\hat{x}_k^i)\}^B$ and optimize the generator to minimize the reconstruction cost as follows:
\begin{equation} \label{eq:2}
    \mathcal{L}_{\theta} = \frac{1}{B} \sum_{i=1}^B \|{\hat{c}_k}^i - {c^{\prime}_k}^i\|^2
\end{equation}
Since we enforce the model to exactly reconstruct the intervened latent representation, from the generated samples alone, the generator can only make the distinct and noticeable change corresponding to the factor encoded in the intervened disentangled code after converging. 

Moreover, we want to ensure that manipulations in the disentangled code $c$ translate to semantic changes in the generated samples thus making them interpretable. To this effect the generator should be constrained to map the intervened representations to examples which lie strictly in the true data manifold. Instead we constrain the encoder to reduce the distance between the posterior distribution $q_{\phi_2} (c)$ and the intervened latent distribution $p(c^{\prime})$. To this effect we train an adversarial network $D_\psi$ to distinguish between the original representations $c$ and the intervened representations $c^{\prime}$ by maximizing the following objective:
\begin{equation} \label{eq:3}
    \mathcal{L}_\psi = \mathbb{E}_{p(c^{\prime})} [\log (D_\psi (c^{\prime})) ] + \mathbb{E}_{c \sim q_{\phi_2} (c|x)} [\log (1-D_\psi (c))] 
\end{equation}
The encoder is in turn trained to fool the discriminator by making the approximate posterior distribution of $c$ indistinguishable from the intervened representations $c^{\prime}$ by minimizing the following object:
\begin{equation} \label{eq:4}
    \mathcal{L}_\phi = \mathbb{E}_{c \sim q_{\phi_2} (c|x)} [\log (1-D_\psi (c))] 
\end{equation}
Since the VAE objective constrains the generator to map the representations to the true data distribution, to minimize the reconstruction cost in the image space, the generator now also maps the intervened representations also to the true data distribution. 

We combine the losses in \{\ref{eq:1}, \ref{eq:2}, \ref{eq:4}\} linearly to train the encoder and the generator to minimize the following objective:
\begin{equation}
    \mathcal{L} = \mathcal{L}_{\theta, \phi} + \lambda \mathcal{L}_\phi + \gamma \mathcal{L}_\theta
\end{equation}
The latent discriminator $D_\psi$ is trained jointly with the other networks to maximize the objective $\mathcal{L}_\psi$. For the first few epochs, we train the model to maximize the likelihood with the discriminator loss and only start the interventions after a few epochs of training. 

\begin{table}[t]
\label{tab::dsp}
\caption{Comparisons of the popular disentanglement metrics on the dSprites dataset. A perfect disentanglement corresponds to 1.0 scores. For all scores, \textbf{higher is better}. The scores are averages over 10 runs with different random seeds. The values for the GAN-based models were taken from \cite{infocr}. Our model outperforms all the VAE-based models while performing comparably with the GAN-based models. $\text{DOT-VAE}^1$ corresponds to the model where intervened values are the values for other samples for the same intervened index whereas $\text{DOT-VAE}^2$ corresponds to the intervened values sampled from the prior distribution.}
\begin{center}
\begin{tabular}{|c|c|c|c|c|c|c|}
\hline
Model & FactorVAE & DCI & MIG &BetaVAE & Modularity & Explicitness\\
\hline
$\beta$-VAE (24)   &0.65 $\pm$ .11& 0.18 $\pm$ .10 & 0.11 &  0.83 $\pm$ .30 & 0.82 $\pm$ .03 & 0.81\\
$\beta$-$\text{TCVAE}$ & 0.76 $\pm$ .18 &  0.30 $\pm$ .05 & 0.18 & 0.88 $\pm$ .18 & 0.85 $\pm$ .03 & 0.83 \\
FactorVAE(40)   & 0.75 $\pm$ .12& 0.26 $\pm$ .04 & 0.15 & 0.85 $\pm$ .15 & 0.81 $\pm$ .02 & 0.80 \\
\hline
InfoGAN$^*$     & 0.82 $\pm$ .01 & 0.60 $\pm$ .02 & 0.22& 0.87 $\pm$ .01 &0.94 $\pm$ .01 & 0.82\\
InfoGAN-CR$^*$      & \textbf{0.88 $\pm$ .01} & \textbf{0.71 $\pm$ .01} & 0.37& 0.95 $\pm$ .01 &\textbf{0.96} & 0.85 \\
\hline
DOT-VAE$^1$ (ours)    &     0.77 $\pm$ .12 & 0.66 $\pm$ .06 & \textbf{0.38} & 0.95 $\pm$ .13 & 0.86 $\pm$ .05 &\textbf{ 0.86 }   \\
DOT-VAE$^2$ (ours)    &     0.72 $\pm$ .15 & \textbf{0.72 $\pm$ .06} & 0.34 &  \textbf{0.97 $\pm$ .18} & 0.82 $\pm$ 0.02& 0.84    \\
\hline
\end{tabular}
\end{center}
\label{tab::dsprites}
\end{table}

\begin{table}[t]
\caption{Quantitative comparisons of the disentanglement metrics on the 3DShapes dataset averaged over 10 runs with different random seeds.  The results for other method have been taken from \cite{loca}.  For all metrics, a higher value indicates a more disentangled latent space. As we can see, DOT-VAE (our model) outperforms most of previous models.}
\label{tab::sh}
\begin{center}
\begin{tabular}{|c|c|c|c|c|c|c|}
\hline
Model & FactorVAE & DCI & MIG &BetaVAE & Modularity & Explicitness\\
\hline
$\beta$-VAE(32)    & 0.82 $\pm$ .24 & 0.58 $\pm$ .40 & 0.34 & 0.99 $\pm$ .06 & 0.94 $\pm$ .18& 0.87 \\
$\beta$-TCVAE & 0.83 $\pm$ .20 & 0.68 $\pm$ .46 & 0.27 & \textbf{1.00 $\pm$ .07}&\textbf{ 0.96 $\pm$ .23} & 0.92  \\
FactorVAE(20)   & 0.81 $\pm$ .26 & 0.64 $\pm$ .27 &0.29 & 0.98 $\pm$ .07 & 0.95 $\pm$ .28 & 0.94 \\
\hline
DOT-VAE$^1$ (ours)    &     \textbf{0.95 $\pm$ .18} & \textbf{0.80 $\pm$ .30} & \textbf{0.42} & \textbf{ 1.00 $\pm$ .04} & 0.94 $\pm$ .19 &\textbf{0.95}   \\
\hline
\end{tabular}
\end{center}
\label{tab::3dshapes}
\end{table}

\section{Related Work}
\label{relatedwork}

Various authors have attempted to learn unsupervised disentangled representations using generative models in recent years. SOTA for unsupervised disentanglement learning can be broadly classified into two categories based on the type of generative model used; one via Variational Autoencoders (VAE) \cite{vae,vae2}, and another via Generative Adversarial Networks (GAN) \cite{gan}. 
Techniques that use VAEs have been modifications of the base VAE architecture to further facilitate a structured latent code. The $\beta$-VAE \cite{beta} heavily penalize the KL divergence term thus forcing the learned posterior distribution $q_{\phi}(z|x)$ to be independent like the prior. AnnealedVAE \cite{cci} controls the capacity of the latent encoding to allow for the independent encoding of the factors. 
Factor-VAE \cite{factor} and $\beta$-TCVAE \cite{tc} penalize the total correlation of the aggregated posterior $q_{\phi} (z)$ 
using adversarial and statistical techniques respectively.
DIP-VAE \cite{dip} forces the covariance matrix of the aggregated posterior $q(z)$ to be close to the identity matrix by method of moment matching. Other works improved the performance augmenting the latent space to include the discrete factors \cite{joint,jeong}, and use optimization techniques based on annealing to encode information effectively in the discrete and continuous factors. 





Models based on GANs explicitly condition the generator with a set of independent latent variables $c$ (by concatenation with random noise $z$), and train the generator to generate data which has high mutual information with $c$. The most prominent work is InfoGAN \cite{info} which learns disentangled, semantically meaningful representations by maximizing a lower bound on the intractable mutual information between the conditioning latent variables $c$ and the generated samples $G(z,c)$. 
InfoGAN-CR \cite{infocr} add a contrastive regularizer to the InfoGAN model, which is trained to predict the changes in the latent space given only the pairs of images. \cite{zhu} augment their objective with a similar self-supervised learning task. In \cite{oogan}, the authors add orthogonal regularization to encourage independent representations. In addition, a major difference between the disjoint sets of InfoGAN and our method is that we explicitly learn the conditioning variables $c$, whereas for the GAN-based methods they are pre-determined. Moreover, we do not need to know the number of factors before-hand and instead learn them as we continue training.   

The authors of \cite{suter} introduced the concept of interventions to study the robustness of the learned representations under the Independent Mechanisms (IM) assumption \cite{causal}, while we use the method of interventions while training the model to disentangle common factors from their entangled set. In \cite{highfid} a VAE is used to disentangle representations and then the representations are passed into a GAN to generate high-fidelity images; our approach instead uses both VAEs and GANs for disentanglement. In \cite{hu} the latent space of entangled representations is conditioned with a disentangled code, which is learned in a supervised way using specific attribute discriminators. As far as the authors are aware, we are the first to split the latent dimension into entangled and disentangled in a completely unsupervised way, as well as the first to combine this with interventions, and with incremental learning.

\section{Empirical Evaluation}
For evaluation, we run experiments on three benchmark datasets: two synthetic datasets generated from independent ground truth factors of variation; dSprites \cite{dsprites} and 3DShapes \cite{shapes}. Each datapoint in these datasets can be exactly described using their factors of generation. For a real-world dataset, with unknown factors of variation, we run our model on the CelebA dataset \cite{celeba}. This dataset has a copious amount of noise in each sample, and cannot neatly be described as independent factors.

\textbf{Qualitative Analysis} To demonstrate the disentangling ability of our model, we perform \textit{traversals} across the latent space, and plot the resulting reconstruction, in line with the standard methods for disentanglement. A model has better disentanglement capability if a traversal across the latent space matches with a change in a single generative factor. We demonstrate this both for the synthetic datasets (Fig. \ref{fig:latent_shapes}), as well as the real-world sets (Fig. \ref{fig::celebA}).

\textbf{Quantitative Analysis} We also evaluate the learned representations using three kinds of quantitative metrics as described in \cite{zaidi}. Specifically, we use the FactorVAE\cite{factor}, the BetaVAE\cite{beta}; Mutual Information Gap (MIG)\cite{tc} and Modularity\cite{ridgeway}; and the Disentanglement-Completeness-Informativeness (DCI) \cite{eastwood}  and Explicitness \cite{ridgeway}.  Results for all metrics are found in Table   \ref{tab::dsprites} for dSprites Table \ref{tab::3dshapes} for 3DShapes. We used the implementation from \cite{loca} to calculate all metrics.

\subsection{Results and Discussion}
As we can see in Figure \ref{fig:latent_shapes}, our model successfully disentangles the dSprites and the 3DShapes dataset, with the traversals showing a change in that latent value corresponds to a change in the true factor of variation. Importantly, the "entangled" latent space $z$ encodes no information, and the model was able to, one at a time, disentangle the relevant factors of variation and encode them in $c$. This is in line with our hypothesis for synthetic datasets as the data can be exactly coded as the independent factors of variation.  Our model is able to completely capture all of the factors of variation for dSprites as well as 3DShapes. We believe part of the reason for the success is the iterative, one at a time decomposition, which allows the network itself to discover how many factors there are, contrasted with prior methods that force the network to factorize the data in a given sized latent space. While the latent traversals show near perfect disentanglement, some of the metrics still report a lower value. We believe that this is because the metrics expect each factor to be completely described by a single latent variable. We do not restrict a factor to correspond to be encoded by only one latent variable as doing so amounts to complete unsupervised disentanglement as described in \cite{loca} which has been proved to be impossible. Instead we encode it in a subset of the latent variables in $c$. Thus each dimension of $c$ is encodes only one factor while each factor can be encoded in multiple dimensions of $c$. This is a common issue discussed in the community as elaborated by \cite{eastwood}. 
In Fig. \ref{fig::celebA}, our results for CelebA show that the model can discern the different factors while maintaining a low reconstruction error. 
 
 \subsection{Conclusion}
In this work, we present DOT-VAE, or Disentangling One at a Time VAE, a method of disentangling latent factors of variation in a dataset without any \textit{a priori} knowledge of how many factors the dataset contains.  We demonstrate that DOT-VAE is on par with or outperforms state of the art methods for disentanglement on standard metrics, and generates crisp, clear and interpretable latent traversal reconstructions for qualitative evaluation. Future directions could be to disentangle multiple datasets together using DOT-VAE while ensuring that the model reuses the learned factors and does not catastrophically forget them when new datasets are introduced. 

\end{document}